\documentclass{article}
\usepackage{spconf,amsmath,graphicx,amssymb}
\usepackage{bm}
\usepackage{graphicx}
\usepackage{subfigure}
\usepackage{verbatim}

\title{WQT and DG-YOLO: towards domain generalization in underwater object detection}
\name{Hong Liu\textsuperscript{1,2}, Pinhao Song\textsuperscript{1,2*}, Runwei Ding\textsuperscript{1,2} \thanks{*Corresponding author: pinhaosong@pku.edu.cn. This work is supported by National Natural Science Foundation of China (NSFC, No. U1613209)}}
\address{\textsuperscript{1}Key Laboratory of Machine Perception, Shenzhen Graduate School, Peking University\\
\textsuperscript{2}Peng Cheng Laboratory}
\begin{document}
%\ninept
%
\maketitle
\begin{abstract}
A General Underwater Object Detector (GUOD) should perform well on most of underwater circumstances. However, with limited underwater dataset, conventional object detection methods suffer from domain shift severely. This paper aims to build a GUOD with small underwater dataset with limited types of water quality. First, we propose a data augmentation method Water Quality Transfer (WQT) to increase domain diversity of the original small dataset. Second, for mining the semantic information from data generated by WQT, DG-YOLO is proposed, which consists of three parts: YOLOv3, DIM and IRM penalty. Finally, experiments on original and synthetic URPC2019 dataset prove that WQT+DG-YOLO achieves promising performance of domain generalization in underwater object detection.
\end{abstract}
\begin{keywords}
domain generalization, object detection, underwater, data augmentation, domain invariance
\end{keywords}
\section{Introduction}
\label{sec:intro}
More and more research institutes and scientists consider attaching camera to Autonomous Underwater Vehicles (AUVs) and Remotely Operated Vehicles (ROVs) in order to perform different underwater tasks, such as marine organism capturing, ecological surveillance and biodiversity monitoring. Underwater object detection is an indispensable technology for AUVs to fulfill these tasks.

In application, once a underwater object detector aiming at certain categories have been trained, we hope this detector can be applied in any underwater circumstances. As a result, it is necessary to build a General Underwater Object Detector (GUOD). A GUOD faces three kinds of challenges:

(1) It is much harder to obtain underwater images, and the annotation task usually need experts to accomplish, which is costly. Therefore, labeled dataset of underwater object detection is extremely limited, inevitably leading to overfitting of deep model. Data augmentation aims at solving the problem of lack of data. There are three types of augmentation. First, geometrical transformations (e.g., horizontal flipping, rotation, patch crop \cite{liu2016ssd}, perspective simulation \cite{huang2019faster}) have been proved effective in various fields. Second, cut-Paste-based methods (e.g., randomly cut and paste \cite{dwibedi2017cut}, Mixup \cite{zhang2017mixup}, CutMix \cite{yun2019cutmix}, PSIS \cite{wang2019psis}) help model learn contextual invariance. Third, domain-transfer based methods (e.g., SIN \cite{geirhos2018imagenet}) force model to focus more semantic information.

(2) The contradiction between speed and performance becomes even more critical. A GUOD should be able to work in real time, which is a common requirement in robotics field. However, it is impractical to equip small AUVs with high performance hardware. Some works focus on the speed of deep learning model but keep good control of performance decrease, such as MobileNet \cite{howard2017mobilenets}, SSD \cite{liu2016ssd}, YOLOv3 \cite{redmon2018yolov3}.

(3) Deep model severely suffers from domain shift, but a GUOD should be invariant of water quality, which can not only work well in oceans, but also in lakes and rivers. This can be seen as a kind of domain generalization problem that a model trains on source domains but evaluates on an unseen domain. Some domain adaptation (DA) (e.g., style consistency \cite{rodriguez2019domain}, DA-Faster RCNN \cite{chen2018domain}) and domain generalization (DG) (e.g., JiGEN \cite{carlucci2019domain}, MMD-AAE \cite{li2018domain}, EPi-FCR \cite{li2019episodic}) technologies are proposed before. Nevertheless, most of DG works focus on object recognition and DA works can not directly transplant to DG task, so their effectiveness are not proved in DG object detection task.

This work aims to use small dataset with limited domains to train a GUOD. To handle challenge (1), a new augmentation method Water Quality Transfer (WQT) is proposed to enlarge the dataset and increase domain diversity. To handle challenge (2) and (3), DG-YOLO is proposed to further boost domain invariance of object detection based on a real-time detector YOLOv3. Our method is implemented on Underwater Robot Picking Contest 2019 (URPC2019) dataset, and achieve performance improvement.
\begin{figure}[!htp]
  \centering
  \includegraphics[width=8.7cm,height=3.5cm]{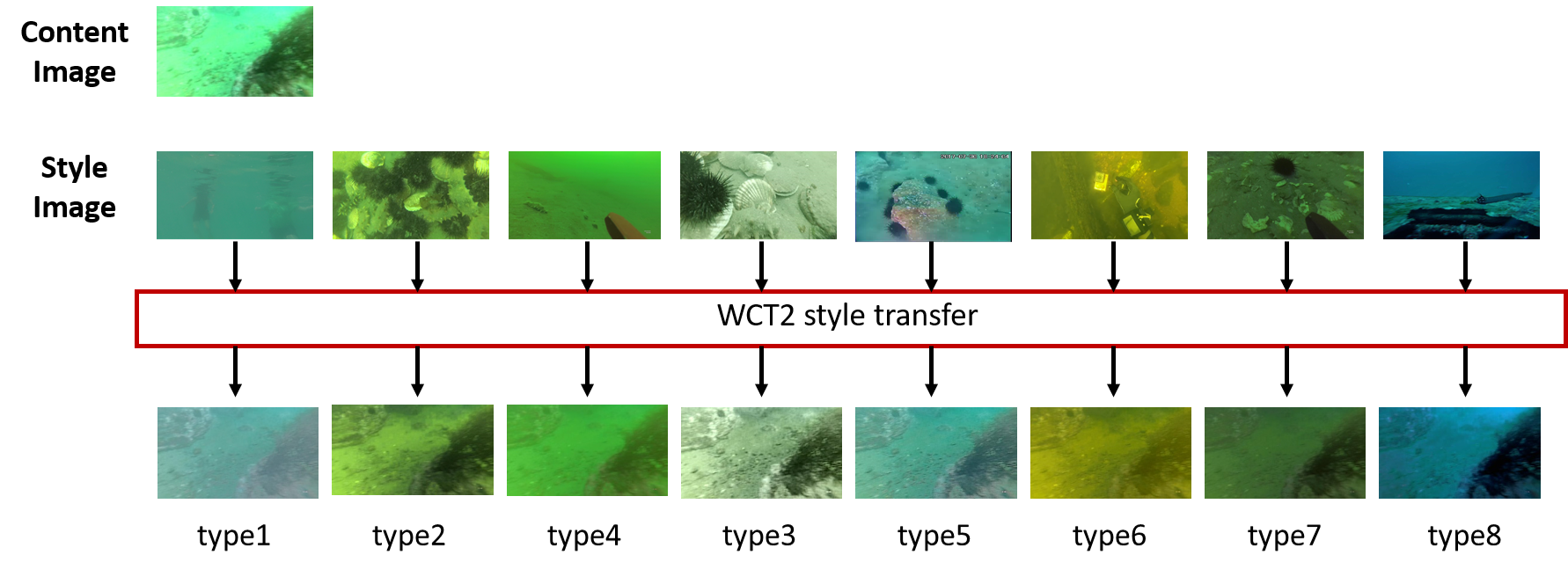}
  \caption{Water Quality Transfer.}
  \label{fig:wqt}
\end{figure}

In summary, our contributions are listed as follows: (1) We propose a new augmentation WQT specially for underwater condition and analyze its effectiveness and reveal its limitations; (2) Based on WQT, DG-YOLO is proposed to further mine the domain-invariant (semantic) information of underwater image, which realizes domain generalization; (3) A lot of experiments and visualization are conducted to prove the effectiveness of our method.

\section{Method}

\subsection{Water Quality Transfer (WQT)}

As Figure \ref{fig:wqt} shows, we select 8 images with different types of water quality, and use $WCT^2$ \cite{yoo2019photorealistic} to transfer URPC dataset to different types of water quality. The content image is from URPC's training set and validation set. In the following section, this seven types of training set are denoted as type1 to type7 and the corresponding validation set are denoted as \emph{val\_type1} to \emph{val\_type7}. As for type8, only the validation set is transferred to obtain \emph{val\_type8} without corresponding training set.
Since model will never train on type8 domain, \emph{val\_type8} is to test the domain generalization capacity of model. 

\subsection{Domain Generalization YOLO (DG-YOLO)}
\label{sec:DG-YOLO}

\textbf{A review of YOLOv3}. Because AUVs with a small processing unit have limited calculation capacity, the real-time detector YOLOv3 \cite{redmon2018yolov3} is a promising choice. YOLOv3 is a one-stage object detector, using Darknet-53 as backbone. Compared with Faster R-CNN \cite{ren2015faster}, YOLOv3 does not use region proposal network. It directly regresses the coordinates of bounding box and class information with a fully convolutional network. YOLOv3 divides an image into $S \times S$ cells, and each cell is responsible for the objects lie in the cell.
The training losses of YOLOv3 consists of the loss of classification $L_{cls}$, the loss of coordination $L_{coord}$, loss of object $L_{obj}$ and loss of no-object $L_{noobj}$:
\begin{equation}\label{eq:yolo-loss}
L_{yolo} = L_{cls}+\lambda_{coord} \cdot L_{coord}+L_{obj}+\lambda_{noobj} \cdot L_{noobj},
\end{equation}
where $\lambda_{coord}$ and $\lambda_{noobj}$ are trade-off parameters.

\begin{figure}[!htp]
  \centering
  \includegraphics[width=8.7cm,height=3.5cm]{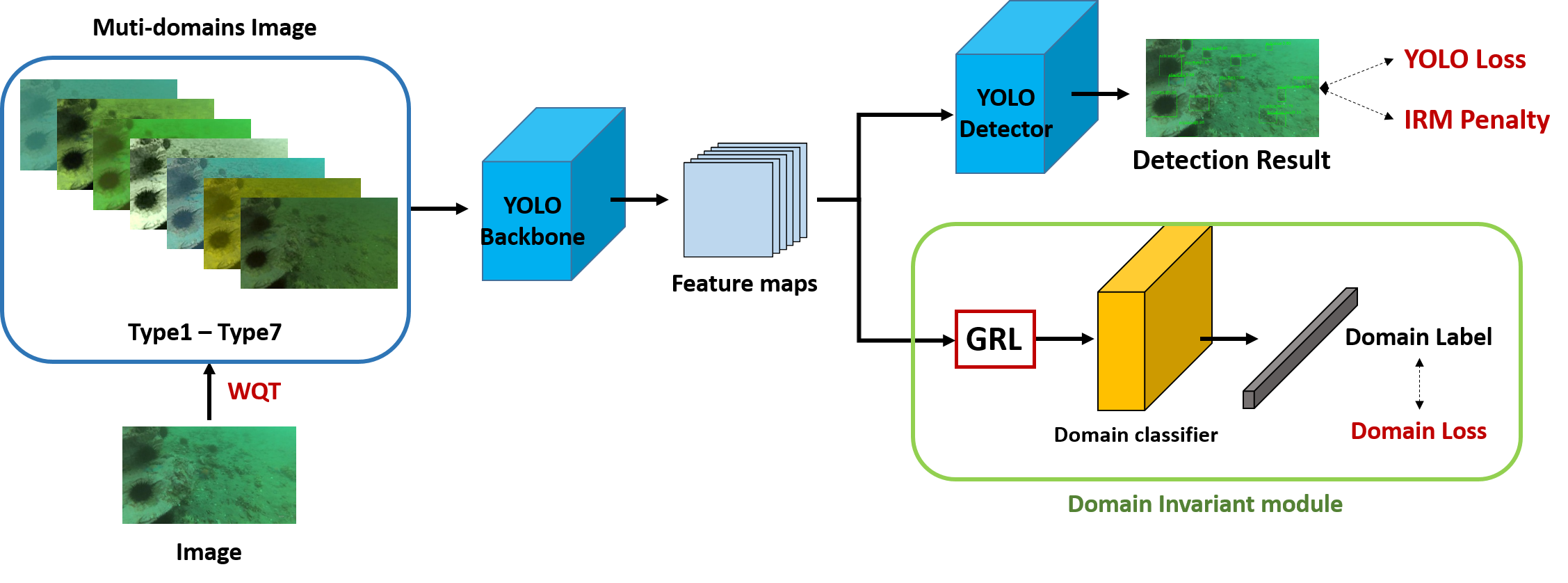}
  \caption{The pipeline of WQT and DG-YOLO. \emph{YOLO BackBone} + \emph{YOLO Detector} is original YOLOv3. Domain label comes from WQT.}
  \label{fig:dg-yolo}
\end{figure}

\textbf{Domain Invariant Module (DIM)}. Since DA and DG have some similarities, we modify the domain classifier proposed by \cite{ganin2015unsupervised} to apply in our DG task. Given a batch of input images as $\{X_1,X_2,...,X_N\}$ from $K$ different source domains, its corresponding domain labels are $\{d_1,d_2,...,d_N\}$, in which $N$ is the number of batch, $d_i\in \mathbb{R}^{K\times 1}$. Denoting that $G$ is a feature extractor and $D$ is a domain classifier, the domain loss $L_{d}$ is defined as follows:

\begin{equation}\label{eq:domain-loss}
L_d = \sum_{i}^{N} l_{CE}(D(G(X_i)),d_i),
\end{equation}
where $l_{CE}$ means categorical cross entropy. In application, domain label comes from WQT, and $K$ is 7 corresponding to 7 types of water quality that WQT synthesizes. Domain loss for data from original dataset is not calculated.

\textbf{IRM Penalty}. 
Inspired by recent study \cite{arjovsky2019invariant}, Invariant Risk Minimization (IRM) help learn an invariant predictor across multiple domain. Given a set of training environments (same meaning as domains) $e \in \mathcal{E}_{tr} $, our final goal is to achieve good performance across a large set of unseen but related environments $\mathcal{E}_{all}$ ($\mathcal{E}_{tr} \in \mathcal{E}_{all}$). However, directly using Empirical Risk Minimization (ERM) \cite{vapnik1992principles} will lead to overfitting on training environment and learn spurious correlation. In order to generalize well on unseen environments, IRM is a better choice to obtain invariance:
\begin{equation}
{min}_{\Phi:\mathcal{X}\sim \mathcal{Y}}\ \sum_{e \in \mathcal{E}_{tr}}R^e(\Phi)+\lambda \cdot ||\nabla_{r|r=1.0}R^e(r\  \cdot \Phi)||^2,
\end{equation}
where $\Phi$ is the entire invariant predictors, $R^e(\Phi)$ is ERM term on environment $e$, $r=1.0$ is a fixed scalar,  $||\nabla_{r|r=1.0}R^e(r\  \cdot \Phi)||^2$ is invariance penalty, and $\lambda \in [0,\infty)$ is a trade-off parameter balancing the ERM term and the invariance penalty.
To apply IRM in YOLOv3, the IRM penalty specially for YOLOv3 is designed as follows:

\begin{footnotesize}
\begin{equation}
  \begin{aligned}
  & Pen_{coord} = \nabla_{r|r=1.0}||\sum_{i=0}^{S^2}\sum_{j=0}^{B}\mathbb{I}^{obj}_{i,j}[(x_i-\hat{x}_i \cdot r)^2+(y_i-\hat{y}_i\cdot r)^2, \\
  & \ \ \ \ \ \ \ \ \ \ +(w_i-\hat{w}_i\cdot r)^2+(h_i-\hat{h}_i\cdot r)^2]||^2, \nonumber
  \end{aligned}  
\end{equation}
\begin{equation}
  Pen_{cls}= \nabla_{r|r=1.0}||\sum_{i=0}^{S^2}\mathbb{I}^{obj}_{i}\sum_{c \in classes} l_{CE}(p_i(c),\sigma(a_i(c) \cdot r))||^2,  \nonumber
\end{equation}
\begin{equation}
  Pen_{obj}=\nabla_{r|r=1.0}||\sum_{i=0}^{S^2}\sum_{j=0}^{B}\mathbb{I}^{obj}_{i,j}l_{CE}(C_i,\sigma(S_i \cdot r))||^2,  \nonumber
\end{equation}
\begin{equation}
  Pen_{noobj}=\nabla_{r|r=1.0}||\sum_{i=0}^{S^2}\sum_{j=0}^{B}\mathbb{I}^{noobj}_{i,j}l_{CE}(C_i,\sigma(S_i \cdot r))||^2, \nonumber
\end{equation}
\end{footnotesize}
\begin{equation}\label{eq:full-penalty}
  P_{IRM} = pen_{coord}+pen_{cls}+pen_{obj}+pen_{noobj},
\end{equation}
where $\mathbb{I}_{i}^{obj}$ denotes if object appears in cell $i$,  $\mathbb{I}_{i,j}^{obj}$ denotes that the $j$th bounding box predictor in cell $i$ is responsible for that prediction, $\sigma$ is sigmoid operation, $\hat{a_i}(c)$ is the score of class $c$ before sigmoid operation, $p_i \in \mathbb{R}^{K \times 1}$ is class label, $\hat{S_i}$ is the score of objects before sigmoid operation, $C_i \in \{0,1\}$ is object label. $\{\hat{x_i},\hat{y_i},\hat{w_i},\hat{h_i}\}$ is the bounding box outputted by YOLOv3, whose corresponding ground truth is $\{x_i,y_i,w_i,h_i\}$. 

% Penalty term is similar to corresponding yolov3's loss. Compared with $L_{coord}$, $Pen_{coord}$ replaces $\hat{x}_{i}$ with $\hat{x}_{i} \cdot r$, and calculating square gradient of $L_{coord}$ to $r$. Compared with $L_{cls}$, $Pen_{cls}$ replaces $\hat{a}_{i}(c)$ with $\hat{a}_{i}(c) \cdot r$, and calculating square gradient of $L_{cls} $ to $r$. Compared with $L_{obj}$ and $L_{noobj}$, $Pen_{obj}$ and $Pen_{obj}$ replaces $\hat{S}_{i}$ with $\hat{S}_{i} \cdot r$, and calculating square gradient of $L_{cls} $ to $r$.
Penalty term is designed based on corresponding losses of YOLOv3. To be specific, $r$ is added to different places of losses. Square gradient of losses to $r$ is the corresponding penalty term.

\textbf{Network overview}.
An overview of our network is shown in Figure \ref{fig:dg-yolo}, we denote it DG-YOLO. Compared to YOLOv3, DIM and IRM penalty are added. In details, the backbone of YOLO darknet-53 can be regarded as a feature extractor. The feature maps extracted from darknet will be fed into Gradient Reversal Layer (GRL) \cite{ganin2015unsupervised} first, which reverses the gradient when backpropagating for adversarial learning. After that, domain classifier distinguish feature maps between domains. With the help of GRL and domain classifier, the backbone will be forced to abandon information of water quality to fool domain classifier. As a result, DG-YOLO can make a prediction depending more on semantic information. Moreover, IRM penalty is calculated simultaneously with YOLO loss. Combining (\ref{eq:yolo-loss}), (\ref{eq:domain-loss}) and (\ref{eq:full-penalty}) ,the total loss of DG-YOLO is:
\begin{equation}\label{eq:total-loss}
L_{total}=L_{yolo}+\lambda_{d} \cdot L_d+\lambda_{p} \cdot P_{IRM},
\end{equation}
$\lambda_{p}$ and $\lambda_{d}$ set to 1 in experiment. In inference stage, because DIM and IRM penalty can be abandoned, DG-YOLO doesn't affect the speed of YOLOv3. It should be emphasized that because domain label comes from WQT, DG-YOLO can not be used alone without WQT.

\section{Experiments and Discussions}
\label{sec:Experiment and discussion}

\subsection{Dataset}
\label{ssec:dataset}
We evaluate WQT and DG-YOLO on a publicly available datasets: URPC2019$\footnote{www.cnurpc.org}$, which consists of 3765 training samples and 942 validation samples over five categories: echinus, starfish, holothurian, scallop and waterweeds. Applying WQT on training set and validation set of URPC2019, we can synthesize \emph{type1-7} for training and \emph{val\_type1-8} for validation. The performance on \emph{val\_type8} represents domain generalization capacity of model. 
 
\subsection{Training details}
\label{ssec:training details}
YOLOv3 and DG-YOLO is trained for 300 epochs and evaluated on original and all synthetic validation sets, with image resizing to 416 $\times$ 416. Models are trained on a Nvidia GTX 1080Ti GPU with PyTorch implementation, setting batch size to 8. Adam algorithm is adopted for optimization and learning rate sets to 0.001, with $\beta_1=0.9$ and $\beta_2=0.999$. IoU, confidence and non-max suppression threshold all set to 0.5. Accumulating gradient is leveraged, which is summing up the gradient and make one step of gradient descent in each two iterations. We do not use any other data augmentation on YOLOv3 and DG-YOLO unless we mention it.

\begin{table*}[!ht]
  \scriptsize
  \caption{The performance of model augmented by different types of water quality. Evaluation is performed on URPC2019 validation set of different types of water quality. All methods are implemented on YOLOv3.}
  \vspace{7pt}
  \centering
  \setlength{\tabcolsep}{0.8mm}
  \begin{tabular}{ccccccccc}
      \hline
       &   &   &  & Evaluation (mAP) &  &  &  &    \\
      \cline{2-9}
      Method & val\_ori & val\_type1 & val\_type2 & val\_type3 & val\_type4 & val\_type5 & val\_type6 & val\_type7   \\
      \hline
      baseline  & 56.45   & 18.72    & 16.83  & 26.57  &10.71  &23.66  &9.38  &29.04   \\
      ori+type1  & 56.66   & 52.39   & 27.66  & 42.07  & 14.25  & 42.79  &20.96  &41.07   \\
      ori+type2  & 56.71   & 18.90   & 51.86  & 39.44  & 24.89  & 34.51  & 6.21  & 45.85   \\
      ori+type3  & 57.78   & 18.01  & 29.96  & 53.10  & 15.63 & 35.07  & 5.68  & 41.20   \\
      ori+type4  & 58.33   & 16.80  & 33.50  & 41.57  & 53.85  & 35.52  & 3.72  & 42.30   \\
      ori+type5  & 57.63   & 35.35  & 30.73  & 42.04  & 20.04  & 53.12  & 19.41  & 42.28   \\
      ori+type6  & 57.19   & 21.64  & 35.63  & 42.19  & 24.37  & 36.04  & 51.22  & 46.15   \\
      ori+type7  & 58.43   & 7.57  & 34.81   & 39.23  & 15.52  & 32.77  & 3.88  & 52.36   \\
      \textbf{Full\_WQT} & \textbf{58.56}   & \textbf{55.93}  & \textbf{53.60}  & \textbf{57.48}  & \textbf{54.95}  & \textbf{56.08}  & \textbf{53.51}  & \textbf{54.29} \\  
      \hline
      ori+rot+flip & 62.53 & 14.81  & 18.29 & 31.36 & 8.89 & 24.95 & 5.34 & 33.18 \\
      \textbf{Full\_WQT+rot+flip} & \textbf{63.83}   & \textbf{60.57}  & \textbf{57.71}  & \textbf{60.38}  & \textbf{58.96}  & \textbf{59.84}  & \textbf{58.43}  & \textbf{60.53} \\
      \hline    
  \end{tabular}
  \label{tab:wqt performance}
\end{table*}

\subsection{Experiments of WQT}
\label{sec:WQT}
In this subsection, we analyze why WQT works. In Table \ref{tab:wqt performance}, \emph{Ori} means original URPC dataset, \emph{baseline} means YOLOv3 is trained only on original dataset, and \emph{ori+type1} means YOLOv3 is trained with original dataset and type1 dataset. \emph{Full\_WQT} means YOLOv3 is trained across type1 to type7. From Table \ref{tab:wqt performance}, we can find three interesting points:

(1) Compared to \emph{baseline}, it can be concluded that every group of augmentation improves the performance in original validation dataset. WQT can be used together with other data augmentation methods to obtain higher performance (last two rows of Table \ref{tab:wqt performance}), which further proves its effectiveness. Besides, there is a phenomenon that WQT also helps model generalize better on other type of water quality in most of the cases. For example, \emph{ori+type7} evaluates on type3 get mAP 39.23\%, 12.66\% higher than \emph{baseline}. 

(2) We believe that there is a correlation between performance and similarity between water qualities. First, we use style loss proposed by \cite{gatys2016image} to represent style distance, and calculate the style distance between different types water quality. We feed style image type1 to type7 into $WCT^2$, and extract the feature maps at certain layers from both encoder and decoder, calculating style loss between any two types and obtaining $H_{dist}$. The result is shown in Table \ref{tab:styleloss}. Second, we take the data from column 3 to 9 (\emph{val\_+type1} to \emph{val\_type7}) and row 2 to 8 (model \emph{ori+type1} to \emph{ori+type7}) in Table \ref{tab:wqt performance}, subtracting each row of this 7 $\times$ 7 matrix to the performance of corresponding type of \emph{baseline}, getting $H_{perf}$. Using Pearson Correlation Coefficient and taking negative, it can be found that the correlation coefficient between style and performance is \emph{0.4634}. From this analysis, it can be inferred that the increase of generalization capacity gaining from WQT is from the similarity between different types of water quality.

(3) To further analyze the finding of (2), model is evaluated on \emph{val\_type8} which is a very different style from type1 to type7. There is no doubt that the WQT-trained model will perform not only better on original dataset, but also across type1 to type7 dataset. However, the model still fails on \emph{val\_type8} (see Table \ref{tab:ablation}), which is far from the requirement of a GUOD. WQT is not enough for domain generalization.

\renewcommand\arraystretch{1.2}
\begin{table}[h]
  \scriptsize
  \caption{Style distance between types of water quality.}
  \vspace{7pt}
  \centering
  \setlength{\tabcolsep}{0.8mm}
  \begin{tabular}{cccccccc}
      \hline
       & type1 & type2 & type3 & type4 & type5 & type6 & type7    \\
      \hline
      type1  & 0   & 0.6281   & 0.1105  & 0.6893  & 0.0495  & 0.7239  & 0.6286  \\
      type2  & 0.6281   & 0   & 0.2860  & 0.0052  & 0.3311  & 0.0077  & 0.0033  \\
      type3  & 0.1105   & 0.2860   & 0  & 0.3435  & 0.0411  & 0.3575  & 0.2977  \\
      type4  & 0.6893   & 0.0052   & 0.3435  & 0  & 0.3747  & 0.0074  & 0.0037  \\
      type5  & 0.0495   & 0.3311   & 0.0411  & 0.3747  & 0  & 0.4024  & 0.3308  \\
      type6  & 0.7239   & 0.0077   & 0.3575  & 0.0074  & 0.4024  & 0  & 0.0094  \\
      type7  & 0.6286   & 0.0033   & 0.2977  & 0.0037  & 0.3308  & 0.0094  & 0  \\
      \hline       
  \end{tabular}
  \label{tab:styleloss}
\end{table}

\renewcommand\arraystretch{1.2}
\begin{table}[!ht]
  \scriptsize
  \caption{Comparisons with other object detector and ablation study of DG-YOLO. IoU, NMS and Conf threshold all set to 0.5. The best checkpoints on \emph{ori} and \emph{val\_type8} are selected respectively for different methods.}
  \vspace{7pt}
  \centering
  \setlength{\tabcolsep}{0.5mm}
  \begin{tabular}{cccccccc}
      \hline
       & & & & val\_type8 (mAP) & & & \\
      \cline{3-8}
      Method & ori & echinus & starfish & holothurian & scallop & waterweeds & ave.     \\
      \hline
      baseline (YOLOv3)  & 56.45 & 53.51   & 7.32   & 11.15  & 9.89  & 0    & 16.37  \\
      WQT-only  & \textbf{58.56} & 60.98   & 17.08   & 33.29  & \textbf{39.02}  & 2.38    & 30.55  \\
      \hline
      Faster-RCNN+FPN  & 58.20 & 29.49   & 5.91   & 9.13  & 1.07  & 10.40    & 11.23  \\
      SSD512  & 56.51 & 26.62   & 14.44  & 18.07  & 1.41  & 14.5   & 15.22  \\
      SSD300  & 50.66 & 27.31   & 14.57   & 13.62  & 3.01  & 2.98    & 12.31  \\
      \textbf{WQT+DG-YOLO} & 54.81 & \textbf{63.84}   & \textbf{27.37}   & \textbf{35.64}  & 36.88  & \textbf{5.11}  & \textbf{33.77}  \\
      \hline
      WQT+DIM & 58.06 & 58.78   & 18.55   & 26.64  & 21.82  & 4.39    & 26.03  \\
      WQT+$P_{IRM}$   & 57.01 & 54.99   & 25.98   & 32.90  & 29.25  & 0   & 30.63  \\
      \hline       
  \end{tabular}
  \label{tab:ablation}
\end{table}

\subsection{Experiments of DG-YOLO}
\label{sec:DG-YOLO experiment}
\textbf{The effectiveness of DG-YOLO}. WQT helps YOLOv3 to learn domain-invariant information, but the model still suffers from domain shift severely. In Table \ref{tab:ablation}, it is shown that DG-YOLO further digs domain-invariant information from data, obtaining 3.21\% mAP improvement on \emph{val\_type8} compared to \emph{WQT-only}. Besides, compared with other object detectors on \emph{val\_type8} performance, DG-YOLO shows its much better domain generalization capacity.

\textbf{Ablation study}. 
The result of ablation study is shown in Table \ref{tab:ablation}. Compared to \emph{WQT-only} on \emph{val\_type8}, \emph{WQT+DIM} has 4.52\% performance decrease and \emph{WQT+$P_{IRM}$} has little improvement. However, \emph{WQT+DG-YOLO} achieves 3.21\% improvement, which suggests only by combining DIM and $P_{IRM}$ can lead to better performance.
\begin{figure}[h]
  \centering
  \includegraphics[width=7cm,height=2cm]{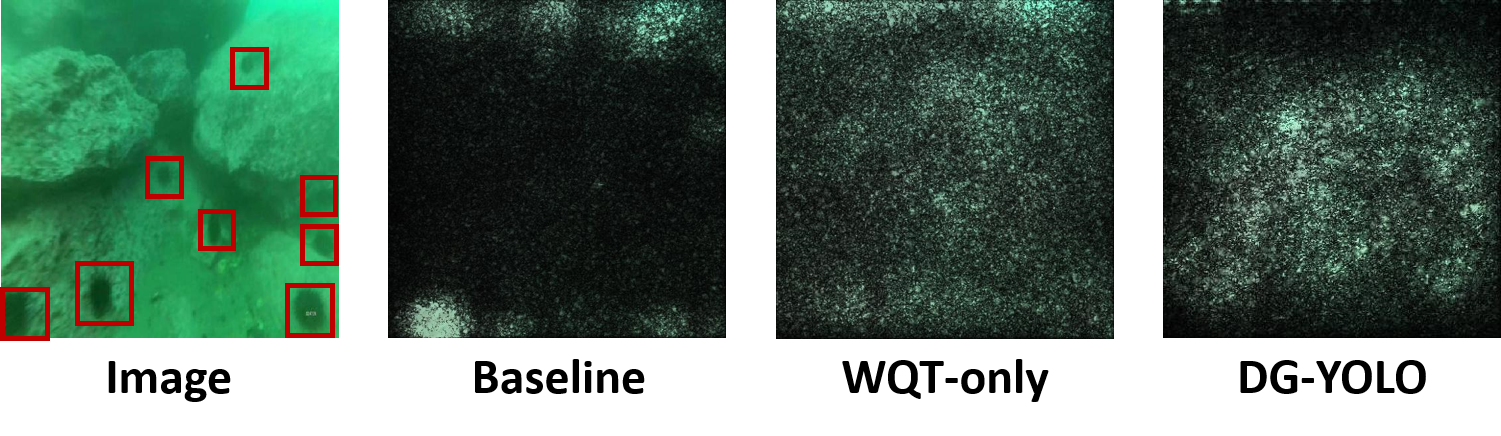}
  \caption{SmoothGrad Visualization on "echinus" samples. Images are from original UPRC2019 dataset.}
  \label{fig:smoothgrad}
\end{figure}

\textbf{Visualization of DG-YOLO}.
One thing that can not be ignored is that there is performance decrease in original validation dataset of \emph{WQT+DG-YOLO}. It is because \emph{WQT-only} is "cheating", learning spurious correlation to make predictions. For example, waterweeds are green in greenish water, but they may turn black in another type of water. Therefore, color of the object is not a domain-invariant information, although it is convenient to use this spurious correlation to achieve good result in just one domain. The performance decrease of DG-YOLO can be interpreted that the model abandons the domain-related information and tries to learn domain invariant information from dataset. We use SmoothGrad \cite{smilkov2017smoothgrad} visualization technique to prove our hypothesis, finding the area that make model to believe there is echinus with probability higher than 95\%. As is shown in Figure \ref{fig:smoothgrad}, \emph{baseline} focuses on the shadow on the top left of image where there is no echinus. The pixel that \emph{WQT-only} focuses is too dispersed, which means \emph{WQT-only} learns spurious correlation. And the pixel DG-YOLO focuses is concentrated and exactly lie on the place where there is echinus. The visualization shows that DG-YOLO learn more semantic information than \emph{baseline} and \emph{WQT-only}.

\section{Conclusion}
This paper propose a data augmentation method WQT and a novel model DG-YOLO to overcome three challenges a GUOD faces: limited data, real-time processing and domain shift. Leveraging $WCT^2$, WQT is intended to increase domain diversity of original dataset. With DIM and IRM penalty, DG-YOLO can further mine semantic information from dataset. Experiments on original and synthetic URPC2019 dataset prove remarkable domain generalization capacity of our method. However, since the performance of DG-YOLO in an unseen domain can still not reach similar level as that in the seen domains, there is still a lot to explore in this field.

%\vfill
%\pagebreak

\bibliographystyle{IEEEbib}
\bibliography{strings,refs}

\begin{thebibliography}{10}

\bibitem{liu2016ssd}
Wei Liu, Dragomir Anguelov, Dumitru Erhan, Christian Szegedy, Scott Reed,
  Cheng-Yang Fu, and Alexander~C Berg,
\newblock ``Ssd: Single shot multibox detector,''
\newblock in {\em European conference on computer vision}. Springer, 2016, pp.
  21--37.

\bibitem{huang2019faster}
Hai Huang, Hao Zhou, Xu~Yang, Lu~Zhang, Lu~Qi, and Ai-Yun Zang,
\newblock ``Faster r-cnn for marine organisms detection and recognition using
  data augmentation,''
\newblock {\em Neurocomputing}, vol. 337, pp. 372--384, 2019.

\bibitem{dwibedi2017cut}
Debidatta Dwibedi, Ishan Misra, and Martial Hebert,
\newblock ``Cut, paste and learn: Surprisingly easy synthesis for instance
  detection,''
\newblock in {\em Proceedings of the IEEE International Conference on Computer
  Vision}, 2017, pp. 1301--1310.

\bibitem{zhang2017mixup}
Hongyi Zhang, Moustapha Cisse, Yann~N Dauphin, and David Lopez-Paz,
\newblock ``mixup: Beyond empirical risk minimization,''
\newblock in {\em International Conference on Learning Representations}, 2018.

\bibitem{yun2019cutmix}
Sangdoo Yun, Dongyoon Han, Seong~Joon Oh, Sanghyuk Chun, Junsuk Choe, and
  Youngjoon Yoo,
\newblock ``Cutmix: Regularization strategy to train strong classifiers with
  localizable features,''
\newblock in {\em Proceedings of the IEEE International Conference on Computer
  Vision}, 2019, pp. 6023--6032.

\bibitem{wang2019psis}
Wang Hao, Wang Qilong, Yang Fan, Zhang Weiqi, and Zuo Wangmeng,
\newblock ``Data augmentation for object detection via progressive and
  selective instance-switching,''
\newblock {\em arXiv preprint arXiv:1906.00358v2}, 2019.

\bibitem{geirhos2018imagenet}
Robert Geirhos, Patricia Rubisch, Claudio Michaelis, Matthias Bethge, Felix~A
  Wichmann, and Wieland Brendel,
\newblock ``Imagenet-trained cnns are biased towards texture; increasing shape
  bias improves accuracy and robustness,''
\newblock in {\em International Conference on Learning Representations}, 2018.

\bibitem{howard2017mobilenets}
Andrew~G Howard, Menglong Zhu, Bo~Chen, Dmitry Kalenichenko, Weijun Wang,
  Tobias Weyand, Marco Andreetto, and Hartwig Adam,
\newblock ``Mobilenets: Efficient convolutional neural networks for mobile
  vision applications,''
\newblock {\em arXiv preprint arXiv:1704.04861}, 2017.

\bibitem{redmon2018yolov3}
Joseph Redmon and Ali Farhadi,
\newblock ``Yolov3: An incremental improvement,''
\newblock {\em arXiv preprint arXiv:1804.02767}, 2018.

\bibitem{rodriguez2019domain}
Adrian~Lopez Rodriguez and Krystian Mikolajczyk,
\newblock ``Domain adaptation for object detection via style consistency,''
\newblock in {\em British Machine Vision Conference}, 2019.

\bibitem{chen2018domain}
Yuhua Chen, Wen Li, Christos Sakaridis, Dengxin Dai, and Luc Van~Gool,
\newblock ``Domain adaptive faster r-cnn for object detection in the wild,''
\newblock in {\em Computer Vision and Pattern Recognition (CVPR)}, 2018.

\bibitem{carlucci2019domain}
Fabio~M Carlucci, Antonio D'Innocente, Silvia Bucci, Barbara Caputo, and
  Tatiana Tommasi,
\newblock ``Domain generalization by solving jigsaw puzzles,''
\newblock in {\em Proceedings of the IEEE Conference on Computer Vision and
  Pattern Recognition}, 2019, pp. 2229--2238.

\bibitem{li2018domain}
Haoliang Li, Sinno Jialin~Pan, Shiqi Wang, and Alex~C Kot,
\newblock ``Domain generalization with adversarial feature learning,''
\newblock in {\em Proceedings of the IEEE Conference on Computer Vision and
  Pattern Recognition}, 2018, pp. 5400--5409.

\bibitem{li2019episodic}
Da~Li, Jianshu Zhang, Yongxin Yang, Cong Liu, Yi-Zhe Song, and Timothy~M
  Hospedales,
\newblock ``Episodic training for domain generalization,''
\newblock in {\em International Conference on Computer Vision (ICCV)}, 2019.

\bibitem{yoo2019photorealistic}
Jaejun Yoo, Youngjung Uh, Sanghyuk Chun, Byeongkyu Kang, and Jung-Woo Ha,
\newblock ``Photorealistic style transfer via wavelet transforms,''
\newblock in {\em International Conference on Computer Vision (ICCV)}, 2019.

\bibitem{ren2015faster}
Shaoqing Ren, Kaiming He, Ross Girshick, and Jian Sun,
\newblock ``Faster r-cnn: Towards real-time object detection with region
  proposal networks,''
\newblock in {\em Advances in neural information processing systems}, 2015, pp.
  91--99.

\bibitem{ganin2015unsupervised}
Yaroslav Ganin and Victor Lempitsky,
\newblock ``Unsupervised domain adaptation by backpropagation,''
\newblock in {\em International Conference on Machine Learning}, 2015, pp.
  1180--1189.

\bibitem{arjovsky2019invariant}
Martin Arjovsky, L{\'e}on Bottou, Ishaan Gulrajani, and David Lopez-Paz,
\newblock ``Invariant risk minimization,''
\newblock {\em arXiv preprint arXiv:1907.02893}, 2019.

\bibitem{vapnik1992principles}
Vladimir Vapnik,
\newblock ``Principles of risk minimization for learning theory,''
\newblock in {\em Advances in neural information processing systems}, 1992, pp.
  831--838.

\bibitem{gatys2016image}
Leon~A Gatys, Alexander~S Ecker, and Matthias Bethge,
\newblock ``Image style transfer using convolutional neural networks,''
\newblock in {\em Proceedings of the IEEE conference on computer vision and
  pattern recognition}, 2016, pp. 2414--2423.

\bibitem{smilkov2017smoothgrad}
Daniel Smilkov, Nikhil Thorat, Been Kim, Fernanda Vi{\'e}gas, and Martin
  Wattenberg,
\newblock ``Smoothgrad: removing noise by adding noise,''
\newblock {\em arXiv preprint arXiv:1706.03825}, 2017.

\end{thebibliography}

\end{document}